\documentclass[sigconf]{acmart}

\usepackage{subfigure}
\usepackage[ruled,linesnumbered]{algorithm2e}
\usepackage{anyfontsize}
\usepackage{bbding}
\usepackage{subfigure}
\usepackage{textcomp}
\usepackage{color}
\pdfoutput=1
\AtBeginDocument{%
  \providecommand\BibTeX{{%
    \normalfont B\kern-0.5em{\scshape i\kern-0.25em b}\kern-0.8em\TeX}}}

\renewcommand\footnotetextcopyrightpermission[1]{} 
\begin{document}

\title{Anomaly Detection on IT Operation Series \\ via Online Matrix Profile}

\author{Shi-Ying Lan}
\authornote{Both authors contributed equally to this research work.}
\email{sylan@stu.xmu.edu.cn}
\author{Run-Qing Chen}
\authornotemark[1]
\email{chenrq1010026261@stu.xmu.edu.cn}
\affiliation{%
  \institution{Xiamen University}
  \city{Xiamen}
  \country{China}
}
\author{Wan-Lei Zhao}
\affiliation{%
  \institution{Xiamen University}
  \city{Xiamen}
  \country{China}}
\email{wlzhao@xmu.edu.cn}

\renewcommand{\shortauthors}{Lan and Chen, et al.}
\renewcommand{\shorttitle}{Anomaly Detection on IT Operation Series via Online Matrix Profile}

\begin{abstract}
Anomaly detection on time series is a fundamental task in monitoring the Key Performance Indicators (KPIs) of IT systems. Many of the existing approaches in the literature show good performance  while requiring a lot of training resources. In this paper, the online matrix profile, which requires no training, is proposed to address this issue. The anomalies are detected by referring to the past subsequence that is the closest to the current one. The distance significance is introduced based on the online matrix profile, which demonstrates a prominent pattern when an anomaly occurs. Another training-free approach spectral residual is integrated into our approach to further enhance the detection accuracy. Moreover, the proposed approach is sped up by at least four times for long time series by the introduced cache strategy. In comparison to the existing approaches, the online matrix profile makes a good trade-off between accuracy and efficiency. More importantly, it is generic to various types of time series in the sense that it works without the constraint from any trained model.
\end{abstract}



\maketitle

\section{Introduction}
In the era of big data, a huge amount of time series data are generated each day from various sources, such as finance, IT security, medical, web services, social media, and geological information systems. Among the huge amount of data, key performance indicators (KPIs) are widely defined in different forms as the major measurement to monitor the performance of a system. For instance, the KPIs in web service are defined as the number of user accesses, the response time of pages, and the page views, etc. The system KPIs which reflect the health status of machines usually include memory usage, and CPU usage, etc. The KPIs status fluctuates abruptly when anything abnormal happens. The detection of such abnormal statuses, which is known as anomaly detection, is critical to maintaining the health of a system. The timely alarm on these anomalies is expected to trigger human intervention or manual diagnosis of the system.

Recently, IT operation analytics is introduced to monitor the IT operation data via artificial intelligence, which is widely known as artificial intelligence for IT operations (AIOps)~\cite{AIOpsChallenge2017,Dang2019}. Anomaly detection on the KPIs data is one of the fundamental tasks in AIOps. Specifically, given a time series $X=\{x_1, \cdots, x_t, \cdots, x_n\}$, anomaly detection aims to judge whether the status at timestamp \textit{t} is an abnormal status that significantly deviates from the majority of the known statuses from timestamp \textit{1} to \textit{t}. Timely detection of anomalies from time series is challenging. First of all, due to the diversity of IT systems that we are monitoring, it is intractable to detect the anomalies by tracing the possible causes of the anomalies. Moreover, the time series are mixed with noises and may demonstrate non-periodic patterns all the way. As a result, the anomalies cannot be detected by simply striking out the non-periodic patterns. Furthermore, there could be several hundreds of KPIs being monitored simultaneously, which requires the detection approach to be generic to different types of data and light-weight.

In recent literature, encouraging performance is reported by several supervised approaches~\cite{Liu2015,Shipmon2017}. However, they are hardly feasible in practice due to two major reasons. Firstly, since anomalies are in rare occurrences, we are lack of true-positives to train a model. Moreover, the annotation of anomalies on a large amount of time series is laborious and error-prone. Secondly, it is not uncommon that the KPI trend drifts when the outside environment changes. This requires the trained model to be updated incrementally. In this case, human intervention is inevitable to decide when the model should be re-trained, which in turn requires further data annotation.

In contrast to supervised approaches, existing statistical approaches require no data annotation. The anomalies are detected as they are outliers compared to the distribution of normal statuses. Traditionally, anomaly detection is built upon status prediction. Typically, the status is judged as abnormal when it is far apart from the predicted value by ARIMA~\cite{Box1976,Salas1980} or Holt-Winters~\cite{Kalekar2004}. However, both ARIMA and Holt-Winters assume the time series values are linear correlated, which is not always satisfied in real-world applications. As a result, the detection results derived from them become very poor as well. Recently, variational auto-encoder (VAE)~\cite{Kingma2013} is adopted to learn the distribution of the normal statuses~\cite{Xu2018, Zhang2019, Park2018, Chen2021}. Since only the major distribution will be encoded by VAE, the statuses are detected as anomalies when they are far away from their decoded statuses. In~\cite{Zhang2019, Park2018}, the long short-term memory (LSTM)~\cite{Hochreiter1997} is integrated to boost the performance of the VAE model.
In recent work~\cite{Chen2021}, spectral residual (SR)~\cite{Hou2007}, which distills the outliers in the frequency domain, is integrated to boost the performance of the VAE-LSTM model. Although superior performance is reported from~\cite{Chen2021}, the learning requires a lot of resources which makes it hard to deal with hundreds of time series simultaneously given the fact that one learned model can only be used for one time series. Similar to supervised approaches, re-training the model periodically is necessary due to the drifting of KPI trends.

For all the aforementioned approaches (including both supervised approaches and trained statistical models), a certain form of training is involved. It is acceptable when there are only a few time series to be processed. The problem becomes intractable in the case that hundreds of time series to be monitored simultaneously, which requires hundreds of trained models to be updated incrementally. In this case, the costs of training and maintaining these models become prohibitively high.

In this paper, matrix profile (MP)~\cite{Yeh2016} which is originally proposed to discover motifs in time series is adopted for anomaly detection. The data structure of MP keeps the distance from a subsequence to its closest subsequence in the time series. The subsequence is viewed as an anomaly when its MP value is sufficiently high. Although it sounds plausible, a direct application of MP is doomed to poor performance because it is insensitive to amplitude variations and is in low localization accuracy. In this paper, several modifications are introduced to MP to tailor it as a training-free anomaly detection approach. The modifications over the original matrix profile include
\begin{itemize}
    \item The distance significance (DS) is introduced to judge whether an anomaly occurs in a timestamp. Compared to the original matrix profile, much higher localization accuracy is achieved;
    \item Additionally, a cache strategy is introduced. Namely, only statuses from recent periods in the timeline are considered when we calculate the left matrix profile. On the one hand, we have no need to keep the full time series for all the KPIs, which saves up a lot of memory consumptions in practice. On the other hand, the time complexity of calculating matrix profile has been reduced from $O(n^2)$ to $O(c{\cdot}n)$, where $n$ is the time series length and $c$ is the cache size. According to our observation, the calculation is sped up with little degradation on the performance.
\end{itemize}

To further boost the performance of online matrix profile, another light-weight approach spectral residual that works online is adopted. Similar to statistical approaches, our approach requires no data annotation. Whereas, unlike statistical approaches, no statistical estimation or learning on the big amount of data is required. The detection judgment is made by referring to the nearest subsequence in time series. Since the nearest subsequence can be defined on various forms of time series, our approach turns out to be a generic solution for anomaly detection. Comprehensive experiments have been conducted in comparison to both the supervised and statistical approaches. We show that our approach makes a good trade-off between learning cost and detection performance.

The rest of this paper is organized as follows. In Section~\ref{sec:rela}, we presented the related work about anomaly detection. In Section ~\ref{sec:pre}, the basic knowledge about the matrix profile is reviewed. In Section~\ref{sec:method}, the proposed approach is presented. In Section~\ref{sec:exp}, the effectiveness of our approach is evaluated on two datasets. Finally, we conclude our paper in Section~\ref{sec:conc}.

\section{Related Work}
\label{sec:rela}
In the last two decades, many approaches have been proposed one after another to address this challenging issue. Although superior performance was reported by supervised approaches~\cite{Liu2015,Shipmon2017,Ren2019}, unsupervised approaches are preferred since the training conditions of supervised approaches can hardly be met in practice. For this reason, we mainly review unsupervised approaches in this section. In general, the unsupervised approaches can be categorized into three groups. For the first group of approaches, the detection is carried out based on status prediction. Namely, an anomaly is detected when its status value is far away from the predicted value by either ARIMA~\cite{Box1976,Salas1980} or Holt-Winters~\cite{Kalekar2004}. Due to the low prediction accuracy from ARIMA and Holt-Winters, the detection performance is poor.  Recently, HTM~\cite{Ahmad2017} and long short-term memory (LSTM)~\cite{Hochreiter1997} are adopted to achieve higher prediction accuracy. Unfortunately, the detection performance is subjected to the accuracy of the prediction models, most of which suffer from noise and drifting patterns.

Another category of approaches detects anomalies out of normal statuses based on statistical models. Since anomalies are in rare occurrences and distributed as outliers, most of the statistical models attempt to learn the distribution of normal statuses. The anomalies are detected when they fall in the range of outliers. Several ways have been seen in the literature to learn the status distribution. In One-class SVM~\cite{Perkins2003}, the time series is projected into a set of vectors. The anomaly is interpreted as outliers in the projected space. This approach is vulnerable to anomalies in the training data. Besides One-class SVM, PCA~\cite{Ringberg2007} is also adopted for data projection. However, it is only effective when time series data are linearly correlated. In the recent approach DONUT~\cite{Xu2018}, variational auto-encoder (VAE)~\cite{Kingma2013} is used to reconstruct time series data. In~\cite{Zhang2019,Park2018}, LSTM is integrated into VAE to capture the temporal dependency of time series data. In order to alleviate the interference from the anomalies during the training, spectral residual (SR)~\cite{Hou2007} is further introduced to the VAE-LSTM model in PAD~\cite{Chen2021}. Recently, cycle-consistent GAN is also used for time series reconstruction in~\cite{Geiger2020}. The disadvantage of these neural network-based approaches lies in two aspects. Firstly, the training requires a lot of data and hardware resources although no annotation is needed. Different models should be built for different data. It makes the detection a computationally intensive task when there are hundreds of time series to be monitored simultaneously. Secondly, such kind of training has to be repeated periodically to adapt to the drifting patterns. 

In the recent approach~\cite{Ren2019}, spectral residual (SR)~\cite{Hou2007}, which is originally proposed to detect visual saliency in images, is adopted for anomaly detection. The anomalies are discovered in the frequency domain as they are less frequent signals compared with normal statuses. In SPOT and Drift SPOT~\cite{Siffer2017}, the anomalies are modeled as low likelihood events based on the Extreme Value Theory~\cite{DeHaan2006}. The anomalies are identified when they are above a dynamic threshold. Extra efforts are required in both SPOT and Drift SPOT to search for the appropriate threshold.


For the last category of anomaly detection approaches, the judgment is made based on the distance between normal statuses and anomalies. Typically, the status at each timestamp is represented by a segment of statuses close to it. Based on the representation, the degree of being an anomaly is measured for each status. The status is viewed as an anomaly when the distance of the segment to its \textit{k}-th nearest neighbor~\cite{Ramaswamy2000} or the summation of distances from its top-\textit{k} nearest neighbors~\cite{Angiulli2002} is sufficiently large. Besides the distance to its \textit{k} nearest neighbors, the judgment is also counted on the number of the reverse nearest neighbors of one segment~\cite{Radovanovic2015}. When the segment has fewer reverse nearest neighbors, the status is more likely to be an anomaly. Alternatively, the relative density such as LoF~\cite{Breunig2000} and LoOP~\cite{Kriegel2009} of one segment to its \textit{k}-distance neighborhood is also employed to make the judgment. The status is viewed as an anomaly when its density is much lower than the density of its \textit{k}-distance neighborhood. The decision is made by referring to the close segments in the feature space. However, these approaches are sensitive to the setting of parameter \textit{k} and show high time complexity. 

In this paper, we re-investigate the feasibility of detecting anomalies based on the distance between statuses. Namely, each status is represented by a subsequence close to it in our approach. The left matrix profile (MP)~\cite{Yeh2016,Zhu2017a} is calculated for each status. In the MP structure, the distance to its nearest subsequence is kept for each status. The anomalies are identified when their \textit{distance significance} in MP is sufficiently large. To facilitate our discussion, the basic knowledge about the matrix profile is reviewed in the next section.

\section{Preliminaries}
\label{sec:pre}
\begin{figure}
    \centering
    \includegraphics[width=0.95\linewidth]{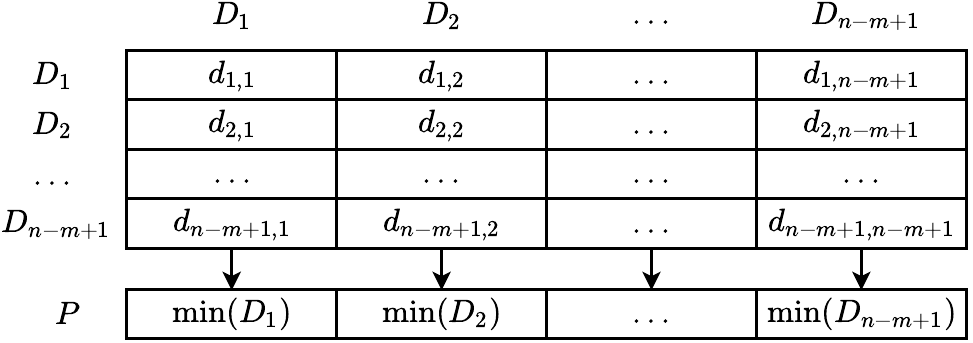}
    \centering
    \caption{An illustration of the relationship between distance profile $D$ and matrix profile $P$.}
    \label{fig:mp2}
\end{figure}
\begin{figure}[!t]
	\centering
	\subfigure[Time series]{\includegraphics[width=0.95\linewidth]{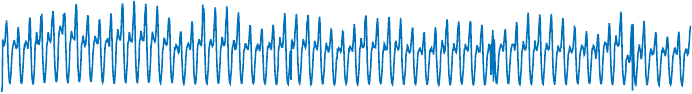}}
	\hspace{0.05cm}
	\subfigure[Matrix profile]{\hspace{8pt}\includegraphics[width=0.92\linewidth]{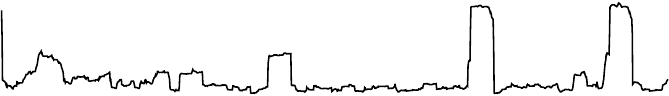}}
    \hspace{0.05cm}
    \subfigure[Left matrix profile]{\hspace{8pt}\includegraphics[width=0.92\linewidth]{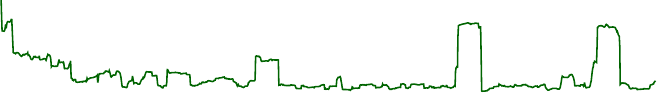}}
	\centering
	\caption{An example of matrix profile and left matrix profile.}
	\label{fig:mp}
\end{figure}

Given a time series $X$, a subsequence $X_{i, m}$ is defined as a continuous interval of length $m$ starting from the position $i$, \textit{i.e.}, $X_{i, m}=\{x_i, x_{i+1},\dots,x_{i+m-1}\}$. The time series $X$ is cut into subsequences with a sliding window. The step size of the sliding window is \textit{1} and the window size is $m$. Given a query subsequence $X_{i, m}$, a vector of its \textit{Euclidean} distance to all subsequences in $X$ is distance profile $D_i$. Formally, $D_i=\{d_{i,1}, d_{i,2}, \dots, d_{i, n-m+1}\}$, where $d_{i,j}$ is the \textit{Euclidean} distance\footnote{It is possible to use other distance measures as long as it is appropriate.} between z-normalized $X_{i,m}$ and z-normalized $X_{j,m}$. Therefore, finding the nearest neighbor (the closest matched subsequence) of $X_{i, m}$ is as easy as minimizing the distance profile $D_i$. To avoid trivial matches~\cite{Mueen2009}, an exclusion zone of $m/2$ before and after the subsequence in the time series is ignored. After minimization, matrix profile $P$ is defined as a vector that keeps the \textit{Euclidean} distances between each subsequence $X_{i,m}$ and its nearest neighbor in the time series $X$, viz., $P=\{\min(D_1), \min(D_2), \dots, \min(D_{n-m+1})\}$. The relationship between distance profile and matrix profile is illustrated in Fig.~\ref{fig:mp2}. Meanwhile, the matrix profile index $I$ is also obtained, which is a vector keeping the position indices for the nearest neighbor of each subsequence.

In order to cancel the drifting patterns and variations in amplitude, the subsequences undergo z-normalization before we calculate the \textit{Euclidean} distance. Given two subsequences $X_{i,m}$ and $X_{j,m}$, the \textit{Euclidean} distance between z-normalized $X_{i,m}$ and z-normalized $X_{j,m}$ is measured as
\begin{equation}
d(X_{i,m}, X_{j,m}) =\sqrt{\left\|\frac{X_{i,m}-\mu_i\textbf{1}}{\sigma_i}-\frac{X_{j,m}-\mu_j\textbf{1}}{\sigma_j}\right\|^2},
\label{eqn:d1}
\end{equation}
where $\mu_i$, $\mu_j$ and $\sigma_i$, $\sigma_j$ are the mean and the standard deviation of two subsequences $X_{i,m}$ and $X_{j,m}$ respectively. Given $n-m+1$ subsequences in $X$, the time complexity of computing $P$ is $O(n^2m)$. If subsequence $X_{i,m}$ is fixed and subsequence $X_{j,m}$ is one of the consecutive subsequences from timestamp \textit{1} to \textit{n}, the distance calculation between $X_{i,m}$ and all the $X_{j,m}$s can be viewed as a series of vector correlation operations. It will be more obvious when Eqn.~\ref{eqn:d1} is re-written in the following form
\begin{equation}
    \begin{aligned}
        d(X_{i,m}, X_{j,m}) = \sqrt{2m\Big(1-\frac{1}{m\sigma_i\sigma_j}(\langle X_{i,m}, X_{j,m} \rangle-m\mu_i\mu_j)\Big)}.
    \end{aligned}
   \label{eqn:d2}
\end{equation}
In Eqn.~\ref{eqn:d2}, the inner-product is undertaken between subsequence $X_{i,m}$ and each $X_{j,m}$, which is a subsequence of $X$ cut out by a sliding window. The most computationally intensive operation in Eqn.~\ref{eqn:d2} is to perform inner-product between $X_{i,m}$ and all consecutive $X_{j,m}$s, which is nothing more than a correlation between subsequence $X_{i,m}$ and sequence $X$. The correlation in Eqn.~\ref{eqn:d2} can be efficiently undertaken in the frequency domain. In scalable time series anytime matrix profile (STAMP)~\cite{Yeh2016}, both $X_{i,m}$ and $X$ are transformed into the frequency domain by Fast Fourier Transform (FFT)~\cite{Loan1992}. The correlation between $X_{i,m}$ and $X$ in the time domain can be converted to inner-product in the frequency domain. In STAMP~\cite{Yeh2016}, the time complexity is reduced to $O(n^2\log n)$. However, STAMP is infeasible in the scenario that time series incrementally grows. In this case, FFT becomes even more costly since the FFT has to be repeated on a time series of growing size whenever a new status arrives. 

Another algorithm called scalable time series ordered-search matrix profile (STOMP)~\cite{Zhu2016} achieves higher speed-up by exploiting the relationship between $\langle X_{i,m},X_{j,m}\rangle$ and $\langle X_{i-1,m},X_{j-1,m}\rangle$. Namely, given inner-product $\langle X_{i-1,m},X_{j-1,m}\rangle$ is known, inner-product $\langle X_{i,m},X_{j,m}\rangle$ is computed conveniently by Eqn.~\ref{eqn:stomp}.
\begin{equation}
\langle X_{i,m},X_{j,m}\rangle = \langle X_{i-1,m},X_{j-1,m}\rangle - x_{i-1}x_{j-1} + x_{i+m-1}x_{j+m-1}.
    \label{eqn:stomp}
\end{equation}
Since inner-product between two subsequences is the most computationally intensive operation in Eqn.~\ref{eqn:d2}, the computation complexity is reduced from $O(m)$ to $O(1)$ with Eqn.~\ref{eqn:stomp}. To compute the matrix profile for a sequence $X$ sized of \textit{n}, the time complexity of the STOMP algorithm can be as low as $O(n^2)$. In a real scenario, the statuses on the right side of status $x_i$ are unknown at timestamp $t_i$. The left matrix profile~\cite{Zhu2017a} instead of the full matrix profile should be calculated. Fig.~\ref{fig:mp}(b)-(c) show the matrix profile and left matrix profile derived from time series in Fig.~\ref{fig:mp}(a).

Essentially, the left matrix profile keeps the distances between each subsequence with its nearest subsequence. The high MP value indicates that the corresponding subsequence is apparently different from the rest of the subsequences. It implies that an anomaly might occur in this subsequence. Therefore, MP provides a new way to address the issue of anomaly detection. Whereas, the direct application of MP in detection will not lead to a decent performance. First of all, a time series grows longer as time goes on. The incremental calculation of the left matrix profile slows down when the series grows longer. Space to hold the time series grows as well. Given that there are thousands of time series to be processed simultaneously, the space complexity could be very high. Moreover, an anomaly could be wrongly judged as normal when its subsequence is close to another subsequence with an anomaly. Furthermore, anomalies exist in subsequences with large or small amplitude in the time series. Anomalies inside subsequences with small amplitude are not visible due to the small MP values. In this paper, all these issues latent in the original MP are carefully considered. The matrix profile has been tailored into online matrix profile (OMP) for anomaly detection, which shows the best trade-off between detection accuracy and efficiency in comparison to most of the existing supervised and unsupervised approaches.

\section{Anomaly Detection by Online Matrix Profile}
\label{sec:method}
In this section, we are going to show how the standard matrix profile is tailored into an online data structure, that is suitable for anomaly detection on KPIs series.
\subsection{Online Matrix Profile}
In anomaly detection, the left matrix profile is preferred over matrix profile since the future statuses on the right are unknown. Moreover, the left matrix profile can only be computed incrementally since the KPIs status arrives one after another. Given the current timestamp \textit{t}, the corresponding subsequence attaching to it is $X_{t-m+1,m}$. The left matrix profile value at timestamp \textit{t} is given as the distance between $X_{t-m+1,m}$ and its closest subsequence on the left of timestamp \textit{t}. To search for the closest subsequence from timestamp \textit{1} to \textit{t-1}, $X_{t-m+1,m}$ is compared with all the past subsequences. The time complexity of doing brute-force comparison is $O(t{\cdot}m)$. The time complexity can be reduced to $O(t)$ by using the STOMP-like strategy, which will be detailed later.

\begin{figure}
\begin{center}
	\subfigure[Time series]
	{\includegraphics[width=0.95\linewidth]{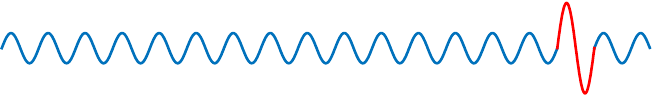}}
	\subfigure[Left matrix profile measured by Eqn.~\ref{eqn:d2}]
	{\hspace{10pt}\includegraphics[width=0.90\linewidth]{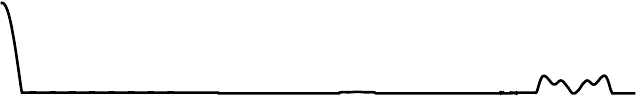}}
	\subfigure[Left matrix profile measured by Eqn.~\ref{eqn:dplus2}]
	{\hspace{10pt}\includegraphics[width=0.90\linewidth]{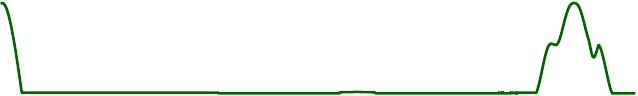}}
\end{center}
	\centering
	\caption{A failed example caused by z-normalization. The time series consists of a \textit{Sine} function with a small amplitude shown in blue. An abnormal subsequence consists of a \textit{Sine} function with a larger amplitude shown in red. After z-normalization, the abnormal subsequence demonstrates a similar amplitude as the rest. The abnormal subsequence becomes insignificant in the left matrix (shown in (b)) derived from this z-normalized time series. Normalization without standard deviation will avoid such failure as shown in (c).}
	\label{fig:sin}
\end{figure}
\begin{figure*}[!t]
	\begin{center}
	\subfigure[Time series]
	{\includegraphics[width=0.95\linewidth]{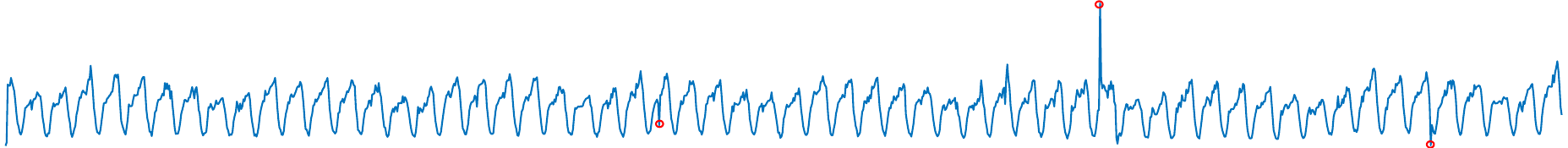}}
	\subfigure[Left matrix profile]
	{\hspace{20pt}\includegraphics[width=0.9\linewidth]{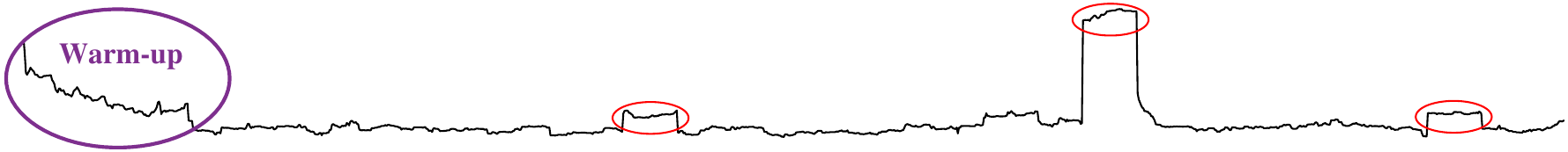}}
	\subfigure[Distance significance]
	{\hspace{20pt}\includegraphics[width=0.9\linewidth]{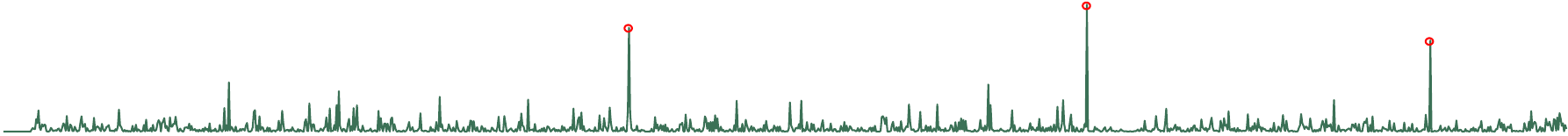}}
\end{center}
    \caption{An example of left matrix profile and distance significance. The time series is shown in blue. The anomalies are spotted by a red circle on each curve. On the left matrix profile curve, there is a subsequence with high matrix profile values at the beginning of the curve. It is called ``warm-up'' stage at the beginning. These high values are mainly due to the limited number of subsequences that the matrix profile could refer to. On the left matrix profile curve, a ``plateau'' segment is observed when an anomaly appears. In contrast, distance significance only shows one high value at the timestamp where the anomaly occurs.}
    \label{fig:ratio}
\end{figure*}
In the standard left matrix profile, the subsequences undergo z-normalization before we compute the \textit{Euclidean} distance between two subsequences. This is to align the mean and standard deviation of all the subsequences to a similar level. After z-normalization, the amplitude variations along the timeline have been canceled. This is helpful to search for motifs~\cite{Yeh2016}, where amplitude variations across different subsequences should be ignored. In anomaly detection, amplitude variation may already indicate a potential anomaly. A typical example is shown in Fig.~\ref{fig:sin}. As shown in the figure, after z-normalization, the abnormal subsequence shown in red demonstrates small values on the left matrix profile curve. To avoid this issue, normalization by the standard deviation is not taken in our implementation. Namely, given the subsequence at timestamp \textit{t} is $X_{t-m+1,m}$, its distance to a subsequence $X_{i,m}$ from within timestamp range $[1, t-1]$ is
\begin{equation}
	\begin{aligned}
	& d(X_{i,m}, X_{t-m+1,m}) =
	\\ & \sqrt{\left\|(X_{i,m}-\mu_i\textbf{1})-(X_{t-m+1,m}-\mu_{t-m+1}\textbf{1})\right\|^2},             
	\end{aligned}
	\label{eqn:dplus1}
\end{equation}
where $\mu_i$ and $\mu_{t-m+1}$ are the mean of $X_{i,m}$ and $X_{t-m+1,m}$ respectively. When a new status $x_t$ comes, we need to calculate the distances between $X_{t-m+1,m}$ and $X_{i,m}$, where $1 \le i \le t-m$. If it is calculated in a brute-force way, the time complexity is $O(t{\cdot}m)$. In order to speed up the calculation, Eqn.~\ref{eqn:dplus1} is re-written as 
\begin{eqnarray}
	\begin{aligned}
		& d(X_{i,m}, X_{t-m+1,m}) =
		\\ & \sqrt{m(\sigma_i^2+\sigma_{t-m+1}^2)-2(\langle X_{i,m}, X_{t-m+1,m}\rangle-m\mu_i\mu_{t-m+1})},
	\end{aligned}
	\label{eqn:dplus2}
\end{eqnarray}
where $\sigma_i$, $\sigma_{t-m+1}$ are the standard deviations of two subsequences, and \textit{m} is the length of one subsequence. Similar to STOMP~\cite{Zhu2016}, the distance calculation between $X_{t-m+1,m}$ and $X_{i,m}$ can be carried out much faster when it is calculated from \textit{1} to \textit{t-m} sequentially. For instance, given inner product $\langle X_{1,m}, X_{t-m,m}\rangle$ is known, the inner product $\langle X_{2,m}, X_{t-m+1,m}\rangle$ is calculated as
\begin{equation}
		\langle X_{2,m}, X_{t-m+1,m}\rangle = \langle X_{1,m}, X_{t-m,m}\rangle - x_1{\cdot}x_{t-m} + x_{m+1}{\cdot}x_t.
		\label{eqn:qt}
\end{equation}
Similarly, all the remaining distances between $X_{t-m+1,m}$ and $X_{i,m}$ can be calculated sequentially from \textit{1} to \textit{t-m}. The time complexity is reduced from $O(t{\cdot}m)$ to $O(t)$. The overall time complexity to compute the left matrix profile for \textit{t} timestamps is $O(t^2)$.

Theoretically speaking, we should refer to all the past normal subsequences to judge whether an incoming subsequence is normal. In practice, most of the normal patterns are in presence when the series is sufficiently long. Therefore, it is no need to keep the statuses of all \textit{t-1} timestamps to be compared with. First of all, it is too costly to keep all the statuses in terms of both computation cost and memory consumption given that there are hundreds of KPIs to be processed. Moreover, the KPI trend drifts as time goes on. It is more reliable to refer to the more recent subsequences. As a consequence, only recent \textit{c} statuses are kept in a cache in our implementation. After the status $x_t$ has been processed, it is pushed into the cache. In the meantime, the oldest status $x_{t-c}$ in the cache is popped out. The time complexity of anomaly detection is reduced to $O(c{\cdot}n)$, given that \textit{n} is the total number of timestamps to be processed. As will be revealed in later experiment, the cache strategy speeds up the detection considerably with only minor degradation on the detection accuracy.
%

%
\subsection{Distance Significance}
In the structure of the left matrix profile, the \textit{Euclidean} distance between each subsequence $X_{i,m}$ and its closest subsequence from within the range of $[1, i-1]$ is kept. The \textit{Euclidean} distance between $X_{i,m}$ and $X_{j,m}$ ($j < i$) is calculated by viewing them as two vectors. Fig.~\ref{fig:ratio}(b) shows the left matrix profile computed for the time series shown in Fig.~\ref{fig:ratio}(a). At the begining, most of the values in the left matrix profile are high due to the lack of reference subsequences. However, the values drop gradually as more and more reference subsequences are available. As shown in Fig.~\ref{fig:ratio}(b), three anomalies are visible from the left matrix profile view. Whereas, they cannot be precisely located. The left matrix profile curve demonstrates high values surrounding the time slot where the anomalies occur. In addition to imprecise localization, the left matrix profile still faces another issue. Since it measures the distance between two subsequences, the left matrix profile demonstrates a larger value when two subsequences are in big amplitude. Accordingly, the left matrix profile value is small when two subsequences are in small amplitude. As a result, anomalies that take place in subsequences with small amplitude are not salient as they are over-dominated by the values from the large amplitude subsequences. In order to address the above issues, distance significance (DS) is introduced.

We focus on judging whether the current status $x_{i+m-1}$ (the last point of the current subsequence $X_{i, m}$) is abnormal when referring to all the subsequences in the cache. Based on the left matrix profile, we already know subsequence $X_{j,m}$ is the closest to $X_{i,m}$. Now the problem is to judge whether $x_{i+m-1}$ is abnormal. To achieve that, the distance significance on timestamp $i+m-1$ is defined as
\begin{equation}
	\hat r_i = \frac{((x_{i+m-1}-\hat \mu_i)-(x_{j+m-1}-\hat \mu_j))^2}{\sum_{w=m-l}^{m-1}((x_{i+w}-\hat \mu_i)-(x_{j+w}-\hat\mu_j))^2},
	\label{eqn:rhat}
\end{equation}
where $\hat \mu_i$ and $\hat \mu_j$ are the means of the last $l$ statuses of $X_{i,m}$ and $X_{j,m}$ respectively. Eqn.~\ref{eqn:rhat} measures the significance that the last status $x_{i+m-1}$ comparing to a strip of distance between $X_{i,m}$ and $X_{j,m}$. If $x_{i+m-1}$ is abnormal, $\hat{r_i}$ is unexpectedly high. It is invariant to amplitude changes since it is a distance ratio relative to its neighboring statuses. The anomaly located in the subsequence with small amplitude is expected to demonstrate big $\hat r_i$ as its distance is in contrast to a summation of small distances. Fig.~\ref{fig:ratio}(c) shows the distance significance computed for the time series shown in Fig.~\ref{fig:ratio}(a). Compared to the left matrix profile (Fig.~\ref{fig:ratio}(b)), the anomalies become prominent even they hold small left matrix profile values. In Eqn.~\ref{eqn:rhat}, \textit{l} is a parameter and is expected to be the same as \textit{m}. In practice, it is possible that there are more than one anomaly within $X_{i,m}$. As a result, \textit{l} is recommended to be smaller than \textit{m} when \textit{m} is a large value, otherwise value $\hat r_i$ is small even anomaly occurs.

Once the distance significance $\hat r_i$ is ready, the detection of anomalies along the time series is as easy as judging whether $\hat r_i$ is above a threshold $\tau$
\begin{equation}
	y_i\ = \begin{cases}
		1, & \hat r_i > \tau \\
		0, & \text{otherwise.}
	\end{cases}
	\label{eqn:y}
\end{equation}
%
\begin{algorithm}[t]
    \caption{Anomaly Detection by Online Matrix Profile}
    \label{alg:OMP}
    \LinesNumbered
    \KwIn{Time series $X$ of cache length $c$, subsequence length $m$, a new status $x_t$ at timestamp $t$, threshold $\tau$, distance threshold $n$}
    \KwOut{Left matrix profile $P$ and its associated left matrix profile index $I$, detection result $y_{t-m+1}$.}
    Initialize distance profile $D_{1\times(c-m)}$ with zero \\
	Push $x_t$ into $X$ \\
	$\mu_{t-m+1} \leftarrow \text{Mean}(X_{t-m+1,m}), \sigma_{t-m+1} \leftarrow \text{Std}(X_{t-m+1,m})$ \\
    \For{$v=1:c-m$}{
        Calculate $\langle X_{t-c+v,m}, X_{t-m+1,m} \rangle$ with Eqn.~\ref{eqn:qt} \\
        Calculate $d(X_{t-c+v,m}, X_{t-m+1,m})$ with Eqn.~\ref{eqn:dplus2}  \\
		$D[v] \leftarrow d(X_{t-c+v,m}, X_{t-m+1,m})$ \\
    }
    Find the minimum value $p_{\text{min}}$ and its associated index $i_{\text{min}}$ of $D$ \\
	Push $p_{\text{min}},i_{\text{min}}$ into $P, I$  \\
	Pop out $x_{t-c}, p_{t-c},i_{t-c}$ from $X,P, I$ \\
	Calculate $\hat r_{t-m+1}$ with Eqn.~\ref{eqn:rhat} \\
	$\tau_d \leftarrow \text{Mean}(P_{t-m+1,m}) + n \times \text{Std}(P_{t-m+1,m})$ \\
	\eIf{$y_{i_{\rm{min}}}$ is $1$ or ($\hat r_{t-m+1} <= \tau$ and $p_{\text{min}} > \tau_d$)}{
			$y_{t-m+1} \leftarrow \text{SR}(X_{t-m+1,m})$ \\
	}{
		\eIf{$\hat r_{t-m+1} > \tau$}{
			$y_{t-m+1} \leftarrow 1$
		}{$y_{t-m+1} \leftarrow 0$}
	}
\end{algorithm}
%
\subsection{Integration with Spectral Residual Analysis}
The detection judgment made by the left matrix profile is unreliable in two cases. In the first case, one anomaly may demonstrate similar pattern as the previous one. Both the matrix profile and distance significance of the current one turn out to be low. As a consequence, such kind of repetitive anomaly cannot be detected by relying on matrix profile alone. As an exception for distance significance, its value is neccessarily high when there are several neighboring timestamps with relative high matrix profile values. In this case, the possible anomaly could be overlooked by the distance significance. In order to detect the anomalies in the above two cases, spectral residual analysis (SR)~\cite{Ren2019} is adopted to make the judgment when either of above two cases occur. SR is compatible with online matrix profile as both of them are free of training and run in an online fashion.


The complete anomaly detection by online matrix profile (OMP) is presented in Algorithm \ref{alg:OMP}. When a new status $x_t$ comes at timestamp $t$, it is appended at the end of the cache (\textit{Line 2}). The subsequence $X_{t-m+1,m}=\{x_{t-m+1}, x_{t-m+2},\dots,x_t\}$ for timestamp $t$ is produced. The mean $\mu$ and the standard deviation $\sigma$ of subsequence $X_{t-m+1,m}$ are calculated (\textit{Line 3}). Thereafter, its distances to all the subsequences from within the cache are calculated. All these distance values are kept as distance profile $D$ (\textit{Lines 4--8}). The minimum value $p_{\text{min}}$ and its associated index $i_{\text{min}}$ of $D$ are the left matrix profile value and the left matrix profile index of $X_{t-m+1,m}$. They are appended to the end of $P$ and $I$ (\textit{Lines 9--10}). The oldest status $x_{t-c}$ in the cache is popped out from the cache (\textit{Line 11}). The distance significance $\hat r_t$ from timestamp $t$ is then calculated (\textit{Line 12}). Since the nearest subsequence $X_{i_{\text{min}}}$ is found and the distance significance $\hat r_t$ is calculated, we can check whether unreasonable match happens. The detection judgment for status $x_t$ is left to SR~\cite{Ren2019} module if the last status $x_{i_{\text{min}+m-1}}$ of $X_{i_{\text{min}}}$ is an anomaly or the left matrix profile $p_{\text{min}}$ is higher than the distance threshold $\tau_d$ while the distance significance $\hat r_t$ is lower than the threshold $\tau$. The distance threshold $\tau_d$ is a dynamic threshold that calculated by the mean and the standard deviation of $P_{t-m+1,m}$. Otherwise, the judgment is made by checking whether $\hat r_t$ is above the threshold $\tau$ (\textit{Lines 13--21}).

\section{Experiments}
\label{sec:exp}
\begin{table}
	\caption{The brief statistics about two evaluation datasets}
	\renewcommand\tabcolsep{3pt}
	\label{tab:data}
	\begin{center}
		\begin{tabular}{|c|rrrc|} \hline
			\textbf{Dataset}  &\textbf{\#Series} &\textbf{\#Timestamps} &\textbf{\#Anomalies} &\textbf{Granularity}\\
			\hline\hline
			\textbf{KPI}   & 29    & 5,922,913 & 134,114 (2.26\%) & Minute \\
			\textbf{Yahoo} & 367   & 572,966   & 3,896 (0.68\%)   & Hour \\ \hline
		\end{tabular}
	\end{center}
\end{table}
\subsection{Datasets and Evaluation Protocol}
Two popular evaluation benchmarks, \textbf{KPI}~\cite{AIOpsChallenge2017} and \textbf{Yahoo}~\cite{YahooLabs2015} are adopted in our experiments. The brief information about these two datasets is summarized in Tab.~\ref{tab:data}. The \textbf{KPI} dataset is released by AIOps Challenge Competition. The KPIs series are collected from various Internet Companies, such as Sogou, eBay, and Alibaba, etc. All the time series are on minute-level. All the anomalies are annotated. The \textbf{Yahoo} dataset is released by Yahoo Lab. The dataset consists of real-world and synthetic time series. The real-world series are collected from the real traffic of Yahoo services. The synthetic data are time series with varying trends, noise, and seasonality. Both the real and synthetic series are on hour-level.

\begin{figure}[t]
	\centering
	\includegraphics[width=0.95\linewidth]{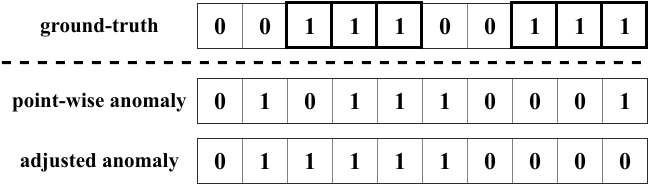}
	\centering
	\caption{The adjustment strategy used in the evaluation.}
	\label{fig:eval}
\end{figure}

Following the convention in the literature~\cite{Xu2018, Ren2019,Chen2021}, precision, recall and $\text{F}_1$-score are used in our evaluation. They are defined in Eqn.~\ref{eqn:precision}, Eqn.~\ref{eqn:recall} and Eqn.~\ref{eqn:f1}, where \#TP, \#FP and \#FN are the number of true positive, false positive and false negative, respectively. Usually, operators are more concerned about whether an anomaly is able to be successfully detected within an acceptable delay in actual application. Therefore, following with the evaluation metrics in AIOps Challenge Competition~\cite{AIOpsChallenge2017}, the detection is viewed as true-positive as long as an anomaly segment is detected in its first $q$ timestamps, where $q$ is a constant. As illustrated in Fig.~\ref{fig:eval}, there are two anomaly segments in the time series shown in the first row, where \textit{0} represents normal status while \textit{1} represents abnormal status. The second row is the detection result. If the allowed delay is one, $i.e.$, $q=1$, the first anomaly segment is adjusted to true while the second one is adjusted to false by applying the adjustment strategy. The detection result after the adjustment is shown on the third row. In the following experiment, we set the delay $q=7$ and $q=3$ for minute-level and hour-level time series, respectively. In addition, both the total and average CPU execution time on the testing data of both datasets are evaluated to compare the efficiency of different approaches.
\begin{equation}
	\label{eqn:precision}
	\text{precision} = \frac{\text{\#TP}}{\text{\#TP}+\text{\#FP}}
\end{equation}
\begin{equation}
	\label{eqn:recall}
	\text{recall} = \frac{\text{\#TP}}{\text{\#TP}+\text{\#FN}}
\end{equation}
\begin{equation}
	\label{eqn:f1}
	\text{F}_\text{1}\text{-score} = \frac{2\times \text{precision} \times \text{recall}}{\text{precision}+\text{recall}}
\end{equation}

Time series in both datasets are under the same pre-processing steps. Following~\cite{Chen2021}, the missing data in both \textbf{KPI} and \textbf{Yahoo} dataset are filled with values that are interpolated with the statuses of the adjacent timestamps or the same time slot from the adjacent periods. For those approaches that require training, the first half of each time series in \textbf{Yahoo} is treated as the training data, while the second half is treated as the testing data. For \textbf{KPI} dataset, we follow the default partition over the training and testing data provided by the benchmark. In our approach, the cache strategy is adopted. The time span of our cache is set to \textit{10} days for both the datasets. Namely, the cache size $c$ in terms of timestamps is set to \textit{14,400} and \textit{240} for the \textbf{KPI} and \textbf{Yahoo} datasets respectively. The sliding window size \textit{m} is set to \textit{2,880} on the \textbf{KPI} dataset, \textit{48} on the \textbf{Yahoo} dataset. The parameter \textit{l} is set to \textit{30} on the \textbf{KPI} dataset, \textit{48} on the \textbf{Yahoo} dataset. The threshold $\tau$ in Eqn.~\ref{eqn:y} is set to \textit{0.37} on the \textbf{KPI}, \textit{0.35} on the \textbf{Yahoo}. The parameter \textit{n} that is to calculate the dynamic distance threshold is set to \textit{1} on the \textbf{KPI}, and \textit{3} on the \textbf{Yahoo}. The above settings are fixed for all our experiments unless otherwise specified.

In this section, we first investigate the effectiveness of each scheme we proposed to enhance matrix profile via an ablation analysis. Thereafter, the performance of our online matrix profile (OMP) is studied in comparison to both unsupervised and supervised approaches in the literature. All the experiments are conducted on a PC with an Intel i5-6500 CPU @ \textit{3.2}GHz (\textit{4} cores) and \textit{16}G memory.
\begin{table*}[!t]
	\caption{Ablation study about OMP on \textbf{KPI} and \textbf{Yahoo}. cache: cache strategy, DS: distance significance, SR: spectral residual}
	\label{tab:abl}
	\begin{center}
		\begin{tabular}{|l|cccr|cccr|}
		\hline
		\multicolumn{1}{|c|}{} & \multicolumn{4}{c|}{\textbf{KPI}} & \multicolumn{4}{c|}{\textbf{Yahoo}} \\
		\hline
		Approach & \multicolumn{1}{c}{F$_1$-score} & \multicolumn{1}{c}{Precision} & \multicolumn{1}{c}{Recall} & \multicolumn{1}{c|}{Time (s)} & \multicolumn{1}{c}{F$_1$-score} & \multicolumn{1}{c}{Precision} & \multicolumn{1}{c}{Recall} & \multicolumn{1}{c|}{Time (s)}\\
		\hline \hline
		\textbf{SR}~\cite{Ren2019}        & 0.622 & 0.647 & 0.598 & 1,027.86 & 0.563 & 0.451 & 0.747 & 28.51 \\
		\textbf{MP}       				& 0.525 & 0.424 & 0.687 & 14,309.17 & 0.599 & 0.679 & 0.536 & 29.10\\
		\textbf{MP${}^*$} 				& 0.597 & 0.565 & 0.633 & 14,309.17 & 0.752 & 0.750 & 0.753 & 29.10\\
		\textbf{MP${}^*$+cache} 		& 0.541 & 0.495 & 0.597 & 3,675.88 & 0.752 & 0.710 & 0.799 & 28.25 \\
		\textbf{MP${}^*$+cache+DS} 		& 0.632 & 0.697 & 0.578 & 3,744.71 & 0.790 & 0.878 & 0.718 & 28.25 \\
		\textbf{OMP} 					& \textbf{0.709} & 0.758 & 0.667 & 3,756.59 & \textbf{0.815} & 0.842 & 0.790 & 28.91 \\
		\hline
	\end{tabular}
	\end{center}
\end{table*}
\begin{table}[t]
	\caption{A close observation on DS performance on the A4Benchmark dataset of \textbf{Yahoo}}
	\label{tab:ds}
	\begin{center}
		\begin{tabular}{|l|ccc|}
		\hline
		\multicolumn{1}{|c|}{} & \multicolumn{3}{c|}{\textbf{Yahoo\_A4Benchmark}}\\
		\hline
		Approach & \multicolumn{1}{c}{F$_1$-score} & \multicolumn{1}{c}{Precision} & \multicolumn{1}{c|}{Recall}\\
		\hline \hline
		\textbf{MP${}^*$+cache} 			 & 0.655 & 0.552 & 0.805 \\
		\textbf{MP${}^*$+cache+DS}        	 & \textbf{0.763} & 0.731 & 0.798 \\
		\hline
		\end{tabular}
	\end{center}
\end{table}

\subsection{Ablation Study}
In our ablation analysis, we investigate the effectiveness of the proposed modifications over the original matrix profile. These modifications include the normalization without standard deviation, the adoption of cache (denoted as cache), the distance significance (DS), and the integration of spectral residual (SR). The spectral residual~\cite{Ren2019} is treated as the comparison baseline as it is the most effective training-free approach in the literature. The performance from the original left matrix profile (given as MP) is also presented. Because the original left matrix profile is unable to localize the anomaly precisely, difference operation is additionally applied on the left matrix profile. This is easily undertaken by subtracting the matrix profile value at timestamp \textit{t-1} from that of timestamp \textit{t}. Several runs of our approach under different configurations are pulled out. Left matrix profile without standard deviation normalization is given as MP*. Based on MP*, we investigate the performance gain when cache, distance significance (DS), and spectral residual (SR) are incrementally integrated. The run that is integrated with cache, DS and SR is the standard configuration of our approach and is called online matrix profile (OMP). The detection performance along with the time costs on the two datasets is presented in Tab.~\ref{tab:abl}.

As shown in Tab.~\ref{tab:abl}, the detection performance is considerably better when the standard deviation is not considered during the normalization. This performance improvement is significant on \textbf{Yahoo} where the amplitude variations are more widely observed on the time series. Although \textit{6\%} performance degradation is observed on \textbf{KPI} when the cache strategy is adopted, the detection is \textit{4} times faster. Speed-up is not observable with \textbf{Yahoo} since the time series is much shorter than that of \textbf{KPI}. Considerable performance improvement is observed on both \textbf{KPI} and \textbf{Yahoo} when distance significance is used to make the judgment. Tab.~\ref{tab:ds} further shows the performance of ``MP${}^*$+cache+DS'' run on a subset of \textbf{Yahoo}, where the amplitude variations along a time series are frequently observed. More than \textit{10\%} performance improvement is achieved over ``MP${}^*$+cache''. Compared with matrix profile distance, distance significance is robust to amplitude variations across different time spans. 

The performance of our approach (namely ``MP${}^*$+cache+DS'') outperforms SR when both cache strategy and distance significance are integrated. In particular, our approach performs significantly better than SR on \textbf{Yahoo}, where anomalies are not necessarily presented as signals with  significant fluctuations. Nevertheless, as both approaches detect the anomalies from different perspectives and are training-free, it possible to integrate SR into our approach. In our standard solution, the detection is left to SR when anomalies occur in the closest subsequence of the current subsequence or when the timestamp shows high matrix profile value while low distance significance. As shown in ``OMP'' run in Tab.~\ref{tab:abl}, \textit{7\%} and \textit{2\%} improvements are observed on \textbf{KPI} and \textbf{Yahoo} respectively when SR is integrated into online matrix profile. In the following experiments, the online matrix profile (OMP), which is integrated with cache strategy, distance significance and spectral residual, is given as the standard configuration of our approach.

\subsection{Anomaly Detection in Comparison to State-of-the-arts}
\begin{figure}[!b]
		\begin{center}
		\subfigure[Non-periodic time series]
		{\includegraphics[width=0.95\linewidth]{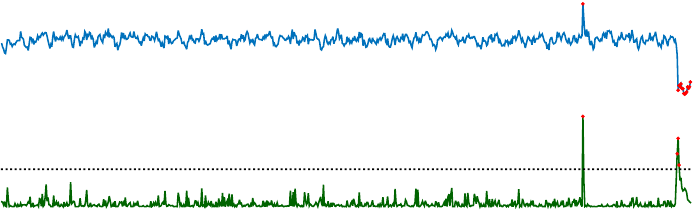}}
		\subfigure[Periodic time series with varying amplitude]
		{\includegraphics[width=0.95\linewidth]{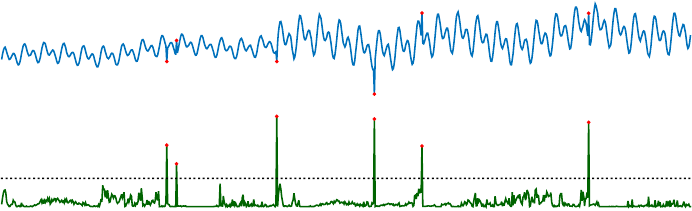}}
		\subfigure[Periodic time series with a repetitive anomaly]
		{\includegraphics[width=0.95\linewidth]{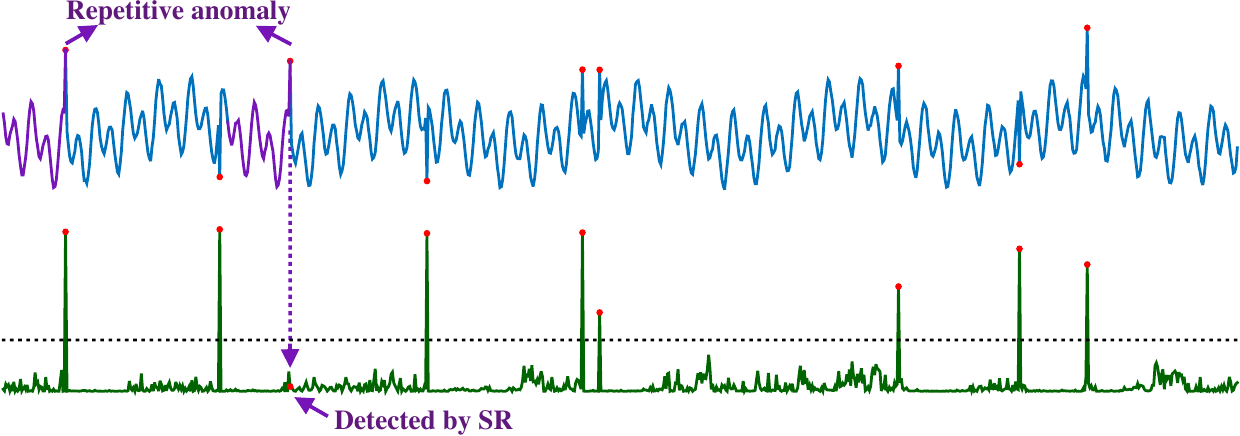}}
		\subfigure[Time series with anomalies that are not easily detectable in the early stage]
		{\includegraphics[width=0.95\linewidth]{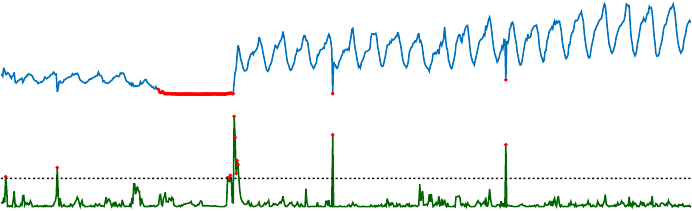}}
	\end{center}
		\caption{Examples of the detection result of OMP on the testing data of \textbf{Yahoo}. The first row and the second row in (a)-(d) are the time series and their distance significance respectively. Each time series is annotated with the ground-truth label. The detection result is shown on the distance significance. The dash line on each distance significance curve indicates the threshold $\tau=0.35$ used in the detection.}
		\label{fig:det}
\end{figure}
\begin{table*}[!t]
	\centering
	\caption{Comparison with the state-of-the-art approaches on testing data. The neural network-based approaches are marked with `\ddag', TM/T: the average time cost to process one timestamp}
	\renewcommand\tabcolsep{4pt}
	\begin{tabular}{|l|c|cccrr|cccrr|}
		\hline
		\multicolumn{2}{|c|}{} & \multicolumn{5}{c|}{\textbf{KPI}} & \multicolumn{5}{c|}{\textbf{Yahoo}} \\
		\hline
		Approach & Online & \multicolumn{1}{c}{F$_1$-score} & \multicolumn{1}{c}{Precision} & \multicolumn{1}{c}{Recall} & \multicolumn{1}{c}{Time (s)} & \multicolumn{1}{c|}{TM/T (ms)} & \multicolumn{1}{c}{F$_1$-score} & \multicolumn{1}{c}{Precision} & \multicolumn{1}{c}{Recall} & \multicolumn{1}{c}{Time (s)} & \multicolumn{1}{c|}{TM/T (ms)} \\
		\hline \hline
		\textbf{SPOT}~\cite{Siffer2017}   & $\surd$ & 0.217 & 0.786 & 0.126 & 1,628.02 & 0.53 & 0.338 & 0.269 & 0.454 & 2,672.62 & 9.33 \\
		\textbf{DSPOT}~\cite{Siffer2017}  & $\surd$ & 0.521 & 0.623 & 0.447 & 1,848.66 & 0.60 & 0.316 & 0.241 & 0.458 & 282.97 & 0.99 \\
		\textbf{SR}~\cite{Ren2019}        & $\surd$ & 0.622 & 0.647 & 0.598 & 1,027.86 & 0.33 & 0.563 & 0.451 & 0.747 & 28.51 & 0.10 \\
		\textbf{${}^\ddag$SR-CNN}~\cite{Ren2019}   & & \textbf{0.771} & 0.797 & 0.747  & 41,438.63 & 13.48 & 0.652 & 0.816 & 0.542 & 336.39 & 1.17 \\	
		\textbf{${}^\ddag$DONUT}~\cite{Xu2018}     & & 0.595 & 0.735 & 0.500   & - 	 & - & 0.501 & 0.669 & 0.401 & - & - \\
		\textbf{${}^\ddag$VAE}~\cite{Kingma2013}   & & 0.685 & 0.725 & 0.648 & 81.70   & 0.03 & 0.642 & 0.773 & 0.549 & 7.87 & 0.03 \\
		\textbf{${}^\ddag$PAD}~\cite{Chen2021} 	  & & 0.739 & 0.839 & 0.660 & 430.30  & 0.14 & 0.755 & 0.837 & 0.688 & 24.08 & 0.08 \\
		\textbf{${}^\ddag$Online-VAE}~\cite{Kingma2013}	  & $\surd$ & 0.686 & 0.716 & 0.657 & 123.05 & 0.04 & 0.541 & 0.694 & 0.443 & 10.92 & 0.04 \\
		\textbf{${}^\ddag$Online-PAD}~\cite{Chen2021} 	  & $\surd$ & 0.731 & 0.806 & 0.669 & 1,560.59 & 0.51 & 0.681 & 0.711 & 0.653 & 59.73 & 0.21 \\
		\hline
		\textbf{OMP} 					  & $\surd$ & 0.709 & 0.758 & 0.667 & 3,756.59 & 1.20 & \textbf{0.815} & 0.842 & 0.790 & 28.90 & 0.10\\
		\hline
	\end{tabular}
	\label{tab:detection}
\end{table*}
\begin{table}
	\caption{The training cost of SR-CNN~\cite{Ren2019}, VAE~\cite{Kingma2013} and PAD~\cite{Chen2021}}
	\label{tab:trntm}
	\begin{center}
		\begin{tabular}{|l|r|r|}
		\hline
		Approach & \multicolumn{1}{c|}{\textbf{KPI} (s)} & \multicolumn{1}{c|}{\textbf{Yahoo} (s)} \\
		\hline \hline
		\textbf{SR-CNN}~\cite{Ren2019}			 & 37,390.82 & 1,415.75 \\
		\textbf{VAE}~\cite{Kingma2013} 			 & 37,412.59 & 1,432.68 \\
		\textbf{PAD}~\cite{Chen2021}        	 & 387,691.11 & 14,535.43 \\
		\hline
		\end{tabular}
	\end{center}
\end{table}
The performance of our OMP is compared with the representative time series anomaly detection approaches in the literature. They include the conventional approaches such as SR~\cite{Ren2019}, SPOT~\cite{Siffer2017}, and DSPOT~\cite{Siffer2017} and the neural network-based approaches such as VAE~\cite{Kingma2013}, DONUT~\cite{Xu2018}, SR-CNN~\cite{Ren2019}, and PAD~\cite{Chen2021}. For VAE and PAD, we also report their performance when their models are updated incrementally, which are given as Online-VAE and Online-PAD. Please note that the same offline training as their offline version for both Online-VAE and Online-PAD is necessary. The $\text{F}_\text{1}$-score, Precision and Recall along with the total and average CPU execution time for all the approaches are reported. For the neural network-based approaches, only the inference time is reported in Tab.~\ref{tab:detection}. The training time costs of these approaches are reported in Tab.~\ref{tab:trntm}. The DONUT approach is a variant of VAE, which shows a similar execution time as VAE. 

The detection results from all the considered approaches are presented in Tab.~\ref{tab:detection}. As shown in the table, OMP achieves the best performance on \textbf{Yahoo} and is only slightly inferior to PAD and Online-PAD on \textbf{KPI}. PAD shows the best performance on \textbf{KPI} under an online environment. However, considerable performance degradation is observed for Online-VAE and Online-PAD when they are tested on \textbf{Yahoo}. Moreover, both of them require a lot of extra time for training. As shown in Tab.~\ref{tab:trntm}, the training cost for PAD could be very high due to its high model complexity. Notice that although SR-CNN shows the best performance on \textbf{KPI}. A large amount of extra time series data is required to train a CNN model to seek a more sophisticated decision rule. It is unrealistic in real-world applications since clean time series are required to support the training. The time cost and the average time cost to process one timestamp are shown in the last two columns of Tab.~\ref{tab:detection} respectively. As shown in the table, most of the approaches could fulfill the detection in real-time. While OMP is the only approach of model-free that shows good performance on two datasets.


The detection samples from OMP on \textbf{Yahoo} are shown in Fig.~\ref{fig:det}. As seen from Fig.~\ref{fig:det}(a)-(b), OMP is able to tackle with periodic and non-periodic time series or time series with varying amplitude. In Fig.~\ref{fig:det}(c), the third anomaly (count from left to right) shows a similar pattern as the first one on the time series. Online matrix profile with distance significance alone fails to identify the third anomaly because both its distance and distance significance to the first anomaly are very small. In this case, SR becomes complementary to the distance significance as it does not rely on the previous subsequences. The detection is, therefore, left to SR. As shown in the figure, the third anomaly on the time series is successfully detected by SR. Fig.~\ref{fig:det}(d) shows a typical case where OMP fails. As seen from the figure, only the last point of the long lasting anomaly is detected since the signals deteriorate into anomalies gradually. Overall, OMP achieves the best trade-off between efficiency and detection accuracy. It is more appealing in real-world KPIs series monitoring as it is an online approach and is able to process multiple time series simultaneously without the constraint from any trained model.

\section{Conclusion}
\label{sec:conc}
We have presented a simple but effective approach, namely online matrix profile for anomaly detection on IT operation time series. The approach is tailored from the matrix profile that is originally used to discover the motifs in time series. With the introduction of distance significance, the anomaly detection becomes invariant to amplitude variations. Meanwhile, it is able to localize the anomalies with considerably higher accuracy than the original matrix profile. Furthermore, a cache strategy is introduced to accelerate the detection. On the one hand, the detection only needs to refer to the most recent subsequences, which is more robust against the drifting trend in KPIs. On the other hand, the time complexity is reduced considerably with only little performance degradation when it processes long time series. Compared with most of the existing anomaly detection approaches, online matrix profile is training-free while achieving competitive performance with the state-of-the-art training-based approaches.

\bibliographystyle{ACM-Reference-Format}
\bibliography{aiops}
\end{document}